\documentclass[10pt,twocolumn,letterpaper]{article}

\usepackage{iccv}             % To produce the CAMERA-READY version
\usepackage{times}
\usepackage{graphicx}
\usepackage{amsmath}
\usepackage{amssymb}
\usepackage{booktabs}

\usepackage{url}            % simple URL typesetting

\usepackage{soul}
\usepackage{color, colortbl}
\usepackage{relsize}
\usepackage{multirow}

\usepackage{multicol}
\usepackage{enumitem}
\usepackage{pifont}

\usepackage[table]{xcolor}
\usepackage{subcaption}
\usepackage[export]{adjustbox}
\usepackage{cellspace, tabularx}

\usepackage[skip=-5pt]{caption}
\usepackage{amsmath}
\usepackage{siunitx}

\usepackage{floatrow}
\usepackage{subcaption} 
\usepackage{color, colortbl}
\definecolor{LightCyan}{rgb}{0.88,1,1}
\definecolor{Gray}{gray}{0.9}

\makeatletter
\renewcommand{\paragraph}{%
  \@startsection{paragraph}{4}%
  {\z@}{1.0em}{-1em}%
  {\normalfont\normalsize\bfseries}%
}
\makeatother

\usepackage{bm}

\usepackage{pifont}
\usepackage{algorithm}% http://ctan.org/pkg/algorithms
\usepackage{algpseudocode}% http://ctan.org/pkg/algorithmicx
\algrenewcommand\algorithmicrequire{\textbf{Input:}}
\algrenewcommand\algorithmicensure{\textbf{Output:}}
\usepackage{enumitem}

\usepackage[english]{babel}
\usepackage[utf8]{inputenc}
\usepackage{amsmath}
\usepackage{bbm}
\usepackage{xcolor}

\DeclareMathOperator*{\argmax}{arg\,max}

\usepackage{amsfonts}
\usepackage{graphicx}
\usepackage{soul}

\definecolor{fig1purple}{RGB}{100,45,100}

\usepackage[pagebackref,breaklinks,colorlinks]{hyperref}

\usepackage[capitalize]{cleveref}
\crefname{section}{Sec.}{Secs.}
\Crefname{section}{Section}{Sections}
\Crefname{table}{Table}{Tables}
\crefname{table}{Tab.}{Tabs.}

\newcommand{\OurDataset}{Coreferenced Image Narratives~}
\newcommand{\OurDatasetShort}{CIN~}

\usepackage{pifont}% http://ctan.org/pkg/pifont
\newcommand{\cmark}{\ding{51}}%
\newcommand{\xmark}{\ding{55}}%
\definecolor{maroon}{HTML}{A9341F}
\definecolor{darkblue}{HTML}{2D2F92}
\definecolor{darkorchid}{HTML}{A4538A} 
\definecolor{bluegreen}{HTML}{00B3B8}
    
\def\iccvPaperID{1725} % *** Enter the ICCV Paper ID here

\iccvfinalcopy % *** Uncomment this line for the final submission

\begin{document}

%%%%%%%%% TITLE - PLEASE UPDATE
\title{Who are you referring to? Coreference resolution in image narrations}

%% FK: I've changed the question in the title: you can't use two "to"s, it either has to be "to whom" or "whom ... to" (preposition splitting). Also, I think "whom" has basically fallen out of use in contemporary English, so I would use "who" in both cases.

% \title{Coreference Resolutions using Weakly Supervised Multimodal Grounding}

\author{\parbox{16cm}{\centering
    {\large Arushi Goel$^1$, Basura Fernando$^{2}$, Frank Keller$^{1}$, and Hakan Bilen$^{1}$}\\
    {\normalsize
    $^1$School of Informatics, University of Edinburgh, UK \\
    $^2$CFAR, IHPC, A*STAR, Singapore.}}
}

\maketitle
%%%%%%%%% ABSTRACT
\begin{abstract}
% \bas{I add some comments in overleaf itself.}
Coreference resolution aims to identify words and phrases which refer to same entity in a text, a core task in natural language processing. 
In this paper, we extend this task to resolving coreferences in long-form narrations of visual scenes. 
First we introduce a new dataset with annotated coreference chains and their bounding boxes, as most existing image-text datasets only contain short sentences without coreferring expressions  or labeled chains. 
We propose a new technique that learns to identify coreference chains using weak supervision, only from image-text pairs and a regularization using prior linguistic knowledge.
Our model yields large performance gains over several strong baselines in resolving coreferences. 
We also show that coreference resolution helps improving grounding narratives in images.
% learns to resolve coreferences in long-form text descriptions, outperforming baselines and a prior model by a significant margin. We also show strong results on the task of weakly supervised visual grounding while resolving coreferences. This is a step towards improving multi-modal linking and contextual reasoning.

% in  dual task of visual phrase grounding in images coupled together with resolving co-references from free form textual descriptions. The task of visual phrase grounding has gained a lot of attention in recent years in computer vision and the task of co-reference resolution is a fundamental problem in the natural language processing community. 
% \bas{should we swap the first and the second sentences?}
% Till now, the extent of resolving phrases and localizing them on images has been done for relatively short and less ambiguous phrases. 
% \bas{Should we also add relatively simple also..Should we motivate the difficult case here by saying "natural spoken language is more ambiguous and complex".}
% Hence, we tackle the challenge of resolving visually and semantically ambiguous phrases without any direct supervision from bounding box annotations in image or co-reference chain linking in text. Solving this challenging problem is a major step towards learning multi-modal linking and contextual reasoning from text as humans do. 

\end{abstract}

%%%%%%%%% BODY TEXT
\section{Introduction}
\label{sec:intro}

% Describe CR in NLP

% Maybe i can add a few more lnes about importance and refer to the problem in Figure 1. 

% Vision + Text for CR 

% \begin{itemize}
%     \item Challenges in CR NLP: domain gap, different cr types
%     \item Visual grounding: Weakly supervised grounding -- Current WS methods do not ground langauge on the images at word level, and due to the annotations’ semantics the grounding boils down mostly to salient objects in images. Due to this limitation, grounding the words in narrations is significantly challenging and cannot be directly used for reference resolution off-the-shelf.
% \end{itemize}

% \hb{it can be good to explain what coreferring expression or mention using the example either in the text and/or caption, use the same term always, e.g. just mention, instead of textual mention or coreferring expression.} \arushi{Got it.}

% \hb{Do we keep the color code for `the woman'? It gives away the answer to our question ;) }\arushi{I think its okay to keep the color code to avoid confusion. Also the point is pretty clear in my opinion. }

\begin{figure}[t]
\begin{center}
\includegraphics[width=0.98\linewidth]{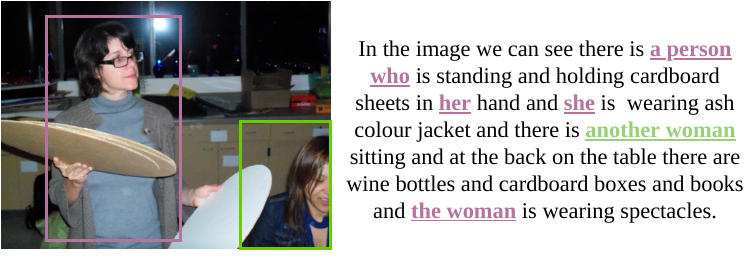}
\vspace{-0.2cm}
\end{center}
\caption{Coreference resolution from an image and narration pair. Each highlighted block of text is referred to as a \textit{mention}. The mentions in the same color corefer to the same entity, belong to the same coreference chain. } 
% For simplicity, we only show \textit{another woman} as an example of a singleton \ie mentions that do not have any corresponding pairs.
% \textit{Figure Source: Localized Narratives \cite{pont2020connecting}.}}
\label{fig:intro-fig}
\end{figure}

Consider the image paired with the long-form description in \Cref{fig:intro-fig}, an example from the Localized Narratives~\cite{pont2020connecting}. 
Can you tell whether \textit{the woman} who is wearing spectacles refers to \textcolor{darkorchid}{\textbf{\textit{a person}}} or \textcolor{green}{\textbf{\textit{another woman}}} in the text?
We are remarkably good in identifying referring expressions (or mentions) and determining which of them corefer to the same entity, a task that we regularly perform when we read text or engage in conversation.
The text-only version of this problem is known as coreference resolution (CR)~\cite{lee2011stanford, lee2017end, sukthanker2020anaphora}, a core task in natural language processing (NLP) with a large literature.
While solving text-only CR requires a very good understanding of the syntactic and semantic properties of the language, the visual version of CR shown in the example also demands understanding of the visual scene.
In our example, a language model has to figure out that \textit{a person} can be a woman, has hands, and correctly match it with \textit{her [hand]} and \textit{the woman}, but not with \textit{another woman}.
However, a language model alone cannot answer whether \textit{the woman} refers to \textit{a person} or \textit{the woman}. 
This can only be disambiguated after visually inspecting which of the two \textit{is wearing spectacles}.

Text-only CR has been a crucial component for a range of NLP applications including question answering~\cite{kwiatkowski2019natural, das2017visual}, sentiment analysis~\cite{cambria2017practical, medhat2014sentiment}, summarization~\cite{gupta2010survey, shi2021neural} and machine translation~\cite{lopez2008statistical, bahdanau2014neural, wu2016google}.
Most text-only CR methods are either rule-based~\cite{lee2011stanford, raghunathan2010multi} using heuristics such as pronoun match or exact match based on part of speech tagging, or are learned on large annotated text datasets from domains such as news text or Wikipedia articles \cite{bengtson2008understanding, lee-etal-2018-higher, lee2017end, joshi2020spanbert}.
State-of-the-art methods~\cite{lee2017end, joshi2020spanbert} fail to resolve coreferences correctly in image narrations for few reasons.
First, CR in image narrations often require image understanding (see \cref{fig:intro-fig}).
Neural networks trained on text datasets~\cite{pradhan2012conll, chen2018preco} suffer from poor transferability and a significant performance drop when applied to image narrations because of domain shift.
Image narrations are unstructured and can be noisy, unlike the well-edited text used during training (such as news or Wikipedia). 
Moreover, standard image-text datasets~\cite{lin2014microsoft, krishna2017visual, changpinyo2021conceptual, plummer2015flickr30k} only contain short descriptions with very few or no cases of coreference, thus, are not suitable for training text-only CR models.

Some prior work have looked at visual CR for specific tasks. 
\cite{ramanathan2014linking} and \cite{rohrbach2017generating} link character mentions in TV shows or movie descriptions to character occurrences in videos. More recently, the Who's Waldo dataset \cite{cui2021s} links person names in the caption to their occurrence in the image.
% Another line of work~\cite{kong2014you} exploits CR to relate texts to 3D scenes. 
% An orthogonal direction of research is to resolve coreference in visual dialog~\cite{kottur2018visual} for developing better question-answering systems for resolving only pronouns. 
However, these methods rely on a limited set of object categories and referring expression types (see \Cref{table:compare} discussed below), require annotated training data and therefore cannot be applied to long-form unconstrained image narrations that include open-world object categories and multiple types of referring expressions such as pronouns (\textit{she}), common nouns (\textit{another woman}), or proper nouns (\textit{Peter}).

%%%% The following is not correct, I'm afraid. An anaphor is essentially a pronoun (well, the term is a bit wider, but in the context of this paper, they are the same):
%\footnote{In the rest of the paper, we use anaphora and %coreference resolution interchangeably.}. 

% \hb{more specific info, what is the unconstrained setting in our problem? different cr types? open vocabulary? long descriptions?} .

In this paper, we look at \emph{the problem of CR in image narrations}, \ie, resolving the coreference of mentions in narrative text that is paired with an image.
As the prior benchmarks in this domain are limited to either a small vocabulary of objects or specific referring expression types, we introduce a new dataset, \OurDataset, \emph{CIN} which augments the rich long-form narrations in the existing Localized Narratives dataset~\cite{pont2020connecting}. 
We add coreference chain annotations and ground each chain by linking it to a bounding box in the corresponding image.

Manually annotating the whole dataset~\cite{pont2020connecting} is expensive, hence these annotations are provided only for evaluation and are not available for training.
To cope with this setting, we propose a weakly supervised CR method that learns to predict coreference chains from only paired image-text data. 
Our key idea is to learn the linking of the mentions to image regions in a joint multi-modal embedding space and use the links to form coreference chains during training. 
To this end, we propose a multimodal pipeline that represents each modality (image regions, text mentions and also mouse traces, additionally provided by \cite{pont2020connecting}) with a modality-specific encoder and then exploit the cross-modal correlations between them to resolve coreference.
Finally, inspired from the rule-based CR~\cite{lee2011stanford}, we incorporate linguistic rules to our learning formulation in a principled way.
We report extensive experiments on \OurDatasetShort and demonstrate that our method not only brings significant improvements in CR but also large gains in weakly supervised narrative grounding, a form of disambiguation that has been underexplored in visual grounding\footnote{Our code and dataset will be made publicly available.}.

 % Contributions
To summarize our contributions, we introduce (1)~the new task of resolving coreferences in multimodal long form textual descriptions (narrations), (2)~a new dataset, \OurDatasetShort, that enables the evaluation of coreference chains in text and the localization of bounding boxes in images, which is provided with multiple baselines and detailed analysis for future work, (3)~a new method that learns to resolve coreferences while jointly grounding them from weak supervision and exploiting linguistic knowledge, (4)~a rigorous experimental evaluation showing significant improvement over the prior work not only in CR but also in weakly supervised grounding of complex phrases in narrative text.

\section{Related Work}
% \hb{don't forget Waldo, other references from the reviews, [1] Wang et.al. Improving weakly supervised visual grounding by contrastive knowledge distillation, Relation-aware Instance Refinement for Weakly Supervised Visual Grounding, CVPR 2021, Improving Weakly Supervised Visual Grounding by Contrastive Knowledge Distillation, CVPR 2021.}

% \hb{It can be good to structure here by using `paragraph'}\arushi{Done.}
% \hb{we can also refer to sup mat for some of the related work}\arushi{Good idea. I can write a more detailed related work for the supp material.}

\paragraph{Text-only CR} in NLP has a long history of rule-based and machine learning-based approaches.  
Early methods~\cite{hobbs1978resolving,raghunathan2010multi} used hand-engineered rules to parse dependency trees, which outperformed all learning-based methods at the time. 
Recently, neural network methods~\cite{wiseman2015learning, wiseman2016learning, clark2016improving, joshi2020spanbert, lee2017end} have achieved significant performance gains. 
The key idea is to identify all mentions in a document using a parser and then learn a distribution over all the possible antecedents for each mention. 
SpanBERT~\cite{joshi2020spanbert} uses a span-based masked prediction objective for pre-training and shows improvements on the downstream task of CR. 
Stolfo~\etal~\cite{stolfo2022simple}, on the other hand, transfer the pretrained representations using rules for CR.
It is worth noting that all these learning-based approaches either require large pretraining data or training data annotated with gold standard (ground-truth) coreference chains, such as OntoNotes~\cite{pradhan2012conll} or PreCo \cite{chen2018preco}.
 
\noindent
\textbf{Visual CR} includes learning to associate people or characters mentioned in the text with images or videos \cite{ramanathan2014linking, rohrbach2017generating, cui2021s}.
Kong~\etal~\cite{kong2014you} exploit CR to relate texts to 3D scenes. 
Another direction is to resolve coreferences in visual dialog~\cite{kottur2018visual} for developing better question-answering systems.
Unlike these works, we focus on learning coreferences from long unconstrained image narrations using weak supervision.
A related group of work~\cite{yu2018mattnet, yu2018rethinking, li2021referring, deng2021transvg} aims to ground phrases in image parts. 
In visual phrase grounding~\cite{yu2018mattnet, liu2019learning, chen2017query, yu2016modeling,deng2021transvg, kamath2021mdetr, li2022grounded}, the main objective is to localize a single image region given a textual query. 
% State of the art methods either use a two-stage \cite{yu2018mattnet, liu2019learning, chen2017query} method by first encoding object regions and then maximizing the similarity with the textual query or a single stage method \cite{deng2021transvg, kamath2021mdetr, li2022grounded}, where the entire image is encoded and then a localization is learned using a standard regression function to find the bounding box coordinates. 
% Single stage approaches are very similar to object detection \cite{ren2015faster, he2017mask} with a conditioned text input. 
These models are trained on visual grounding datasets such as ReferItGame \cite{kazemzadeh2014referitgame}, Flickr30K Entities \cite{plummer2015flickr30k}, or RefCOCO \cite{yu2016modeling}.  
% However, they do not ground all the words in the sentence on the images, and due to short captions, the grounding boils down to mostly salient objects in images. 
However, due to short captions, the grounding of text boils down to mostly salient objects in images. 
In contrast, grounding narrations which aims at capturing all image regions is significantly more challenging and cannot be effectively solved with those prior methods.

% Though these methods can localize a single phrase in the image (achieving up to 80\% accuracy \cite{deng2021transvg}), they fail to disambiguate coreferences (\textit{a lady} and \textit{the back side lady} that refer to different regions/entities in \Cref{fig:intro-fig}), which is the key contribution of our work.
% Few exceptions~\cite{kottur2018visual, rohrbach2017generating, huang2018finding} that can resolve coreference are limited to specific scenarios such as linking pronouns with a fixed set of characters in movies. 

% Limitations of prior image/text methods

\noindent
\textbf{Weakly supervised grounding}, learning to ground from image-text pairs only, has recently been used in \cite{liu2019adaptive, liu2021relation, liu2019knowledge, liu2022entity, wang2021improving} for referring expression grounding. These methods use phrase reconstruction from visual region features as a training signal. 
Other methods~\cite{wang2020maf, gupta2020contrastive, datta2019align2ground} use contrastive learning by creating many negative queries (based on word replacement) or by mining negative image regions for a given query. 
Wang \etal \cite{wang2020maf} is a strong method in this domain, hence we establish it as a baseline in our experiments.
% , there are two other related works.
Liu~\etal~\cite{liu2021relation} parses sentences to scene graphs for capturing visual relation between mentions to improve phrase grounding. 
However, this cannot be directly applied to our task, as parsing scene graphs from narrations is typically very noisy and incomplete.
Wang~\etal~\cite{wang2021improving} aims to learn/predict object class labels from the object detector during training and inference respectively. Due to the open-vocabulary setting in our dataset, we directly rely on predictions from the detector and use them as features to avoid the complexity of open-vocabulary object detection. 
% We compare and show improvements on the MAF model \cite{wang2020maf} as it achieves comparable/better performance on Flickr30k dataset compared to \cite{liu2021relation, wang2021improving}. 
Furthermore, as we show in the experiments that grounding is useful to anchor mentions but it is not sufficient to resolve coreferences without prior linguistic knowledge. Thus, our method also employs contrastive learning but for learning CR from weak supervision. 

% This strategy does not help in
% disambiguating multiple occurences of the same object category in the narration, a major challenge in our dataset unlike existing visual grounding datasets. 

% 

\label{sec:related}

% Phrase Grounding

% Referring Expression Grounding

% Coreference Resoltuion in NLP

\section{\OurDataset} 
\label{sec.dataset}
Our \OurDatasetShort dataset contains 1880 images from the Localized Narratives dataset \cite{pont2020connecting} that come with long-form text descriptions (narrations) and mouse traces. 
These images are originally a subset of the test and validation set of the Flickr30k dataset~\cite{plummer2015flickr30k}.
We annotated this subset with coreference chains and bounding boxes in the image that are linked with the textual coreference chains, and use them only for validation and testing.
Note that we also include singletons (\ie, coreference chains of length one). 
\cref{fig:intro-fig} shows an example image from \OurDatasetShort. 

% \arushi{We can probably change the name of the dataset.}

\noindent
\textbf{Annotation procedure. }
The annotation involved three steps: (1)~marking the mentions (sequences of words) that refer to a localized region in the image, (2)~identifying coreference chains for the marked mentions, including (a)~pronominal words such as \textit{him} or \textit{who} that are used to refer to other mentions, (b)~mentions that refer to the same entity (\eg, \textit{a lady} and \textit{that person}), and (c)~mentions that do not have any links (\eg \textit{another woman}), (3)~drawing bounding boxes in the image for the coreference chains/mentions identified in steps~(1) and~(2). 
We created an annotation interface based on LabelStudio~\cite{labelstudio}, an HTML-based tool that allows us to combine text, image, and bounding box annotation. 
More details are provided in the supplementary material. 
\begin{table}[!ht]
\renewcommand*{\arraystretch}{1.13}

		\resizebox{0.9\textwidth}{!}{
			\begin{tabular}{c | cccc}

			     Dataset & \#noun phrases & \#pronouns & \#coreference chains  & \#bounding boxes   \\ \toprule
			    Flickr30k Entities \cite{plummer2015flickr30k} & 15,252 & \xmark & \xmark & 17,234 \\
			    RefCOCO \cite{yu2016modeling} & 10,668 & \xmark & \xmark & 10,668 \\
			    \textbf{CIN} (Ours) & 19,587 & 1,659 & 3,310 & 21,246\\
				\bottomrule

\end{tabular}}
\caption{Statistics of relevant noun phrases, pronouns, coreference chains and bounding boxes on Flickr30k Entities~\cite{plummer2015flickr30k}, RefCOCO~\cite{yu2016modeling} and \OurDatasetShort.}
\label{table:crg_stats}
\end{table}

\begin{figure}[!ht]
\begin{center}
\includegraphics[width=\linewidth]{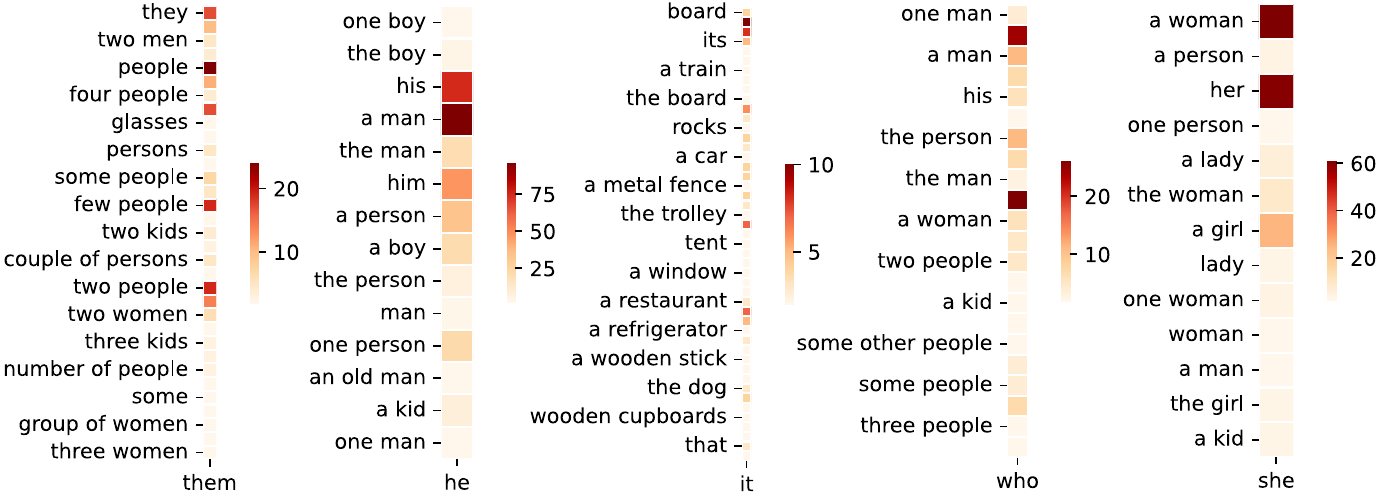}
% \vspace{-0.5cm}
\end{center}
\caption{Numbers of mentions as part of the coreference chain for pronouns \textit{them, he, it, who, she} in \OurDatasetShort. }
% \textit{Figure Source: Localized Narratives \cite{pont2020connecting}.}}
\label{fig:dataset-analysis}
\end{figure}

% \begin{table}[!ht]
% \renewcommand*{\arraystretch}{1.13}

% 		\resizebox{\textwidth}{!}{
% 			\begin{tabular}{ccccc}

% 			     Datasets & &  &  &  \\ \toprule
% 			    Who's Waldo \cite{}  &  & &  &  \\
% 			    RefCOCO \cite{yu2016modeling} &  &  &  &  \\
% 			    \textbf{\OurDataset} (Ours) &  &  &  & \\
% 				\bottomrule

% \end{tabular}}
% \caption{Statistics of relevant noun phrases, pronouns, coreference chains and bounding boxes on Flickr30k Entities \cite{plummer2015flickr30k}, RefCOCO \cite{yu2016modeling} and \OurDataset.}
% \label{table:crg_stats}
% \end{table}

\noindent
\textbf{Dataset statistics.} We split the 1880 images in the dataset into a test and validation set using the pre-defined split of \cite{plummer2015flickr30k}. 
More specifically, we have 1000 images in the test set and 880 images in the validation set. 
It is important to note that the narrations have a lot of first person pronouns such as \textit{I can see \dots}. 
We specifically instruct the annotators to exclude such mentions that are not a part of any coreference chain and at the same time cannot be grounded on the image. We elaborate more on the filtering process for these mentions in the supplementary material.  

Overall, the dataset has 19,587 noun phrase mentions, 1,659 pronouns, 3,310 coreference chains and 21,246 bounding boxes. 
In \Cref{table:crg_stats}, we compare the statistics of \OurDatasetShort with other related datasets. 
In \cref{fig:dataset-analysis}, we show the distribution over the frequency and types of mention such as \textit{a metal fence} or \textit{few people} that are referred to using a particular pronoun (\textit{them, he, it, who} and \textit{she}). 
There is a huge diversity in (1)~the categories of the mentions and (2)~how many times they form a part of the coreference chain. 

\noindent
\textbf{Comparison to existing CR datasets. }
In \Cref{table:compare}, we compare our proposed CIN dataset to other CR datasets. This comparison shows that most of the other datasets are either from a restricted domain (\textit{shopping, indoor scenes, etc.}), have limited mention types referring to either only \textit{people} or \textit{limited object categories}, or do not cover all possible referring expression types such as common nouns (\textit{a person}), proper nouns (\textit{Peter}) and pronouns (\textit{he}). 

\begin{table}[!t]
\renewcommand*{\arraystretch}{1.13}

		\resizebox{\textwidth}{!}{
			\begin{tabular}{c | c| c | c | c }

			     Dataset &  Modality & Domain &  Object categories  & Referring expression types \\ \toprule
			    % VisDial \cite{das2017visual} &  Images & Q/A &   &  \\
                    NYU-RGBD v2 \cite{kong2014you} &  Images & Indoor home scenes & Household objects & Common nouns \\
                    SIMMC 2.0 \cite{guo2022gravl} &  Images & Shopping & Clothing  & Common nouns  \\
			    MPII-MD \cite{rohrbach2017generating} & Videos & Movies &  People & Proper names, Pronouns\\
                    Who's Waldo \cite{cui2021s}  &  Images & WikiMedia &  People  & Proper names \\
			    \textbf{CIN} (Ours) & Images & Open-world & General objects  &  Proper names, Common nouns and Pronouns\\
				\bottomrule

\end{tabular}}
\caption{Comparison to existing datasets. }
% \arushi{Incomplete table, dont know what more columns to add for differentiation. \hb{what about the task or text type? } \arushi{Add comparative images for each dataset in the supp material.} }
\label{table:compare}
\end{table}

\section{Method} 
\label{sec.method}
% !TeX root = main.tex

\subsection{Text-only CR}
\label{sec.crtext}
Given a sentence containing a set of mentions (\ie, referential words or phrases), the task of CR is to identify which mentions refer to the same entity. 
This is fundamentally a clustering problem \cite{sukthanker2020anaphora}.
In this work, we use an off-the-shelf NLP parser \cite{spacy} to obtain the mentions. 
Formally, let $S=\{m_1,m_2,\dots,m_{|S|}\}$ denote a sentence with $|S|$ mentions, where each mention $m$ contains a sequence of words, $\{w_1,w_2,\dots,w_{|m|}\}$.
% The assignment label for each mention pair are denoted as $\{y_1,y_2,\cdot,y_{|s|}\}$ where $y\in {1,2,\dots,C}$ and $C$ is the number of clusters, $C\leq |s|$.
We assign a label $y_{ij}$ to each mention pair $(m_i,m_j)$, which is set to 1 when the pair refers to the same entity, and to $-$1 otherwise.
We wish to learn a compatibility function, a deep network $f$ that scores high if a pair refers to the same entity, and low otherwise.
% Additionaly, we use a mention encoder $e_m$ to encode each mention.

Given a training set $D$ that contains $|D|$ sentences with their corresponding labels, one can learn $f$ by optimizing a binary cross-entropy loss:
\begin{equation}
    \label{eq.ObjectiveSupervisedCR}
    % \min_{\Phi} \frac{2}{|D| (|D|-1)} \sum_{i=1}^{|D|-1}\sum_{j=i+1}^{|D|} y_{ij} \log \sigma(f_{\Phi}(m_i,m_j))
    % \min_{\Phi} \sum_{i=1}^{|D|-1}\sum_{j=i+1}^{|D|} y_{ij} \log \sigma(f_{\Phi}(m_i,m_j)) + (1-y_{ij}) \log (1-\sigma(f_{\Phi}(m_i,m_j)))
    \min_{f} \sum_{S\in D} \sum_{i=0}^{|S|-1}\sum_{j=i+1}^{|S|}  \log  (y_{ij}(\sigma(f(m_i,m_j)))-\frac{1}{2})+\frac{1}{2})
    % (1-y_{ij}) \log  (1-\sigma(f(e_m(m_i),e_m(m_j))
\end{equation} 
where $\sigma$ is the sigmoid function.
Note that prior methods \cite{lee2011stanford,lee2017end, joshi2020spanbert} require large labeled datasets for training and are limited to only a single modality, text. 
These methods typically also combine the learning with fixed rules based on recency and grammatical principles \cite{lee2011stanford}.

\begin{figure*}
\begin{center}
\includegraphics[width=0.8\linewidth]{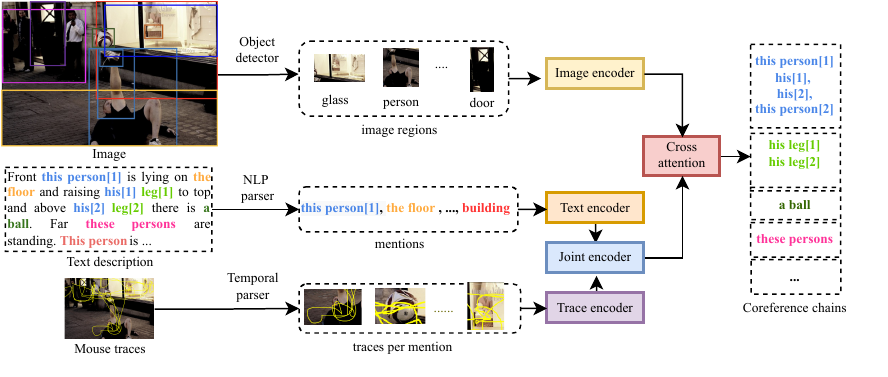}
\vspace{-0.3cm}
\end{center}
\caption{Overview of our pipeline. Our model encodes the image regions obtained from an object detector using the image encoder. We parse text mentions and mouse traces from the sentence description, which are then encoded using a text and trace encoder respectively. Finally, a joint text-trace encoder learns a joint embedding for text and traces. A cross-attention module attends to the words given an image region and then we compute the joint probability of the paired mentions, thus forming coreference chains.}
\label{fig:method-fig}
\end{figure*}

% \noindent
% \textbf{Coreference Resolution in a Multimodal Setting. }
% \subsection{CR for image and text data}
\subsection{CR in image narrations}
\label{sec.crtextimage}
\paragraph{Problem definition.}
Next we extend the text-only CR to image-text data in the absence of coreference labels.
Let $(I,S)$ denote an image-text pair where $S$ describes an image $I$ as illustrated in \Cref{fig:intro-fig}, and assume that coreference labels for mention pairs are not present.
As in \cref{sec.crtext}, our goal is to identify the mentions that refer to the same entity in an image-text pair.
Each image is defined by $|I|$ regions $I=\{r_1,r_2,\dots,r_{|I|}\}$ which are obtained by running the pretrained object detector (trained on the COCO \cite{lin2014microsoft} and Visual Genome \cite{krishna2017visual} dataset) in \cite{ren2015faster} on the image.
Each region $r$ is described by its bounding box coordinates $\bm{b}$, the text embedding for the detected object category $\bm{o}$, and the visual features $\bm{v}$.  
More details are provided in \Cref{sec.model}.

\paragraph{Weak supervision.} We use `weak supervision' to refer to a setting where no coreference label for mention pairs and no grounding of mentions (\ie,~bounding boxes are not linked to phrases in the text) are available.
Moreover, in contrast to the output space of the object detector (a restricted set of object categories), the sentences describing our images come from unconstrained vocabulary.
Hence, an object instance in a sentence can be referred to with a synonym or may not even be present in the object detector vocabulary~\cite{lin2014microsoft, kuznetsova2020open}.
Finally, the object detector can only output \emph{category-level} labels and hence cannot localize object instances based on the more specific \emph{instance-level} descriptions provided by the sentences.
For instance in \Cref{fig:intro-fig}, \textit{a person} and \textit{the woman} both are labeled as \textit{person} by the object detector.

% As mentioned above, we have $|Q|$ textual spans for each sentence and each span $k$ could point to one or multiple image regions. One might think that the object labels and textual spans are strongly related but in practice this is not the case due to two reasons - a) the sentences are noisy and have unconstrained vocabulary so the same object is referred to with a synonym or might not even occur in the object detector vocabulary \cite{} and b), object labels have no notion of context when compared to textual spans (for instance in \Cref{fig:intro-fig}, ``back side lady'' and ``a lady'' both are labeled as ``women'' by the object detector.)

In addition to image and text, we explore the use of an auxiliary modality, mouse trace segments provided in \cite{pont2020connecting}.
Each mouse trace includes a sequence of 2D points over time that relate to a region in the image when describing the scene. 
As the text in Localized Narratives is transcription of the speech of the annotators, the mouse traces are synced with spoken words, which we denote as $T=\{t_1,t_2,\dots,t_{|T|}\}$ where $|T|=|S|$.
These features are stacked with textual features (see \Cref{sec.model}).

% In the special case of the Localized Narratives dataset, we also have mouse trace segments \cite{} for the $N$ words in the sentence denoted as $T \in \{t_0, t_1, \cdots, t_n\}$. 

% \paragraph{Weakly supervised grounding.}
In the weakly supervised setting, the key challenge is to replace the coreference label supervision with an alternative one.
We hypothesize that each mention in a coreferring pair corresponds to (approximately) the same image region, and it is possible to learn a joint image-text space which is sufficiently rich to capture such correlations.
Concretely, let $g(m,r)$ denote an auxiliary function that is instantiated as a deep network and outputs a score for the mention $m$ being located at region $r$ in image $I$.
This grounding score for each mention can be converted into probability values by normalizing them over all regions in the image:
\begin{equation}
    \label{eq.gbar}
    \bar{g}(m,r) = \frac{\exp{(g(m,r))}}{\sum_{r' \in I}\exp{(g(m,r'))}}.
\end{equation}
The compatibility function $f$ can be defined as a sum product of a pair's grounding probabilities over all regions:
\begin{equation}
    \label{eq.f}
    f(m,m') = \sum_{r \in I} \bar{g}(m,r) \bar{g}(m',r).
\end{equation}
In words, mention pairs with similar region correlations yield bigger compatibility scores and are hence more likely to corefer to each other. 
The key idea is that we employ the grounding for mentions as anchors to relate coreferring mentions (\eg, \textit{a person} and \textit{the woman}).
At test time, we compute $f(m,m')$ for all the pairs and threshold them to predict their pairwise coreference labels.
% We also use fixed rules from language to eliminate erroneous labels, which are explained below.

As no ground-truth bounding box for each mention is available for learning the grounding $g$, we pose grounding as a weakly supervised localization task as in \cite{gupta2020contrastive, wang2020maf}.
To this end, we impute the missing bounding boxes by taking the highest scoring region for a given mention $m$ at each training iteration:
\begin{equation}
    \label{eq.imputer}
    r_m = \argmax_{r\in I} g(m,r).
\end{equation}
Then we use $r_m$ as the pseudo-truth to learn $g$ as following:
\begin{equation}
    \label{eq.contrast}
    \min_{g}  \sum_{(I,S)\in D} \sum_{m \in S} - \log \big( \frac{\exp(g(m,r_m))}{\sum_{I' \in D \setminus I} \exp(g(m,r'_{m})) } \big)
\end{equation} 
where $r'_{m}=\argmax_{r\in I'} g(m,r)$ is the highest scoring region in image $I'$ for mention $m$.
For each mention, we treat the highest scoring region in the original image as positive and other highest scoring regions across different images as negatives, and optimize $g$ for discriminating between the two.
% The optimization encourages the model to associate the most probable image region for a mention in the sentence by matching words and the visual features. 
However, as the denominator requires computing $g$ over all training samples at each iteration, which is not computationally feasible, we instead sample the negatives only from the randomly sampled minibatch.

% We aim to optimize the parameters $\Theta = \{\phi, \psi, \theta, \omega \}$ of our model jointly using supervision from positive image-sentence pairs. More specifically, we use contrastive loss learning by treating all the other image regions in the batch as negatives for the positive image-sentence pair. 

% \begin{equation}
%     \min_{\Theta} \frac{1}{\mathcal{B}} \sum_{\mathcal{B}} \frac{1}{|Q|} \sum_{i=1}^{|Q|}  - \text{log} \big( \frac{exp(g(i, r_m))}{\sum_{r^{'} \in \mathcal{B}} exp(g(i, r^{'}_m))} \big) 
%     \label{eq.contrast}
% \end{equation} where $ \mathcal{B}$ are the samples in a minibatch.
% \arushi{I only sum over minibatches in this equation, not the entire dataset. can we change this eq somehow? }
% This enforces the model to learn the most probable image region for a textual span in the sentence by matching words and the visual features. 

\paragraph{Linguistic constraints.} Learning the associations between textual and visual features helps with disambiguating coreferring mentions, especially when mentions contain visually salient and discriminative features. However, resolving coreferences when it comes to pronouns (\eg, \textit{her, their}) or ambiguous phrases (\eg, \textit{one man} or \textit{another man}) remains challenging.
To address such cases, we propose to incorporate a regularizer into the compatibility function $f(m,m')$ based on various linguistic priors.
Concretely, we construct a look-up table for each mention pair $q(m,m')$ based on the following set of rules \cite{lee2011stanford}:

\noindent
\textbf{(a)~Exact String Match.} Two mentions corefer if they exactly match and are noun phrases (not pronouns). 

\noindent
\textbf{(b)~Pronoun Resolution.} Based on the part-of-speech tags for the mentions, we set $q(m,m')$ to 1 if $m$ is a pronoun and $m'$ is the antecedent noun that occurs before the pronoun. 

\noindent
\textbf{(c)~Distance between mentions.} Smaller distance is more likely to indicate coreference since mentions to occur close together if they refer to the same entity.

\noindent
\textbf{(d)~Last word match.} In certain cases, the entire phrases might not match but only the last word of the phrases.

\noindent
\textbf{(e)~Overlap between mentions.} If two mentions have one or more overlapping words, then they are likely to corefer.

% \arushi{Add a figure to show rule matrix for a narration. }

Finally, we include $q(m,m')$ as a regularizer in \cref{{eq.contrast}}:
\begin{equation}
    \label{eq.finalobj}
    \begin{split}
        \min_{g}  \sum_{(I,S)\in D}  \sum_{m \in S} \Big( - \log \big( \frac{\exp(g(m,r_m))}{\sum_{I' \in D \setminus I} \exp(g(m,r'_{m})) } \big) \\
        + \lambda \sum_{m' \in S} ||f(m,m')-q(m,m')||_{F}^{2} \Big)
    \end{split}
\end{equation} 
where $\lambda$ is a scalar weight for the Frobenius norm term.
Note that $f$ is a function of $g$ (see \cref{eq.f}).
We show in \Cref{sec.results} that incorporating this term results in steady and significant improvements in CR performance.

\subsection{Network modules}
\label{sec.model}
Our model (illustrated in \Cref{fig:method-fig}) consists of an image encoder $e_i$ and text encoder $e_t$ to extract visual and linguistic information respectively, and a cross-attention module $a$ for their fusion.
% Formally, the scoring function $g$ can be decomposed as $g(m,r)=a_{(m,r)}(e_t(S),e_i(I))$.
% $g(m,r)=a(e_i(I),e_t(S),e_m(T))$

\vspace{-.3cm}

\paragraph{Image encoder $e_i$} takes in a $d_r$-dimensional vector for each region $r$ that consists of a vector consisting of bounding box coordinates $\bm{b}\in R^{4}$, text embedding for the detected object category $\bm{o}\in R^{d_o}$ and visual features $\bm{v}\in R^{d_v}$.
The regions are extracted from a pretrained object detector~\cite{ren2015faster} for the given image $I$.
The image encoder applies a nonlinear transformation to this vector to obtain a $d$-dimensional embedding for each region $r$.

\vspace{-.3cm}

\paragraph{Text encoder $e_t$} takes in the multiple mentions from a parsed multi-sentence image description $S$ produced by an NLP parser \cite{spacy} and outputs a $d$-dimensional embedding for each word in the parsed mentions.
Note that the parser does not only extract nouns but also pronouns as mentions.
% In particular, $e_t$ first encodes words in each mention and projects them to a $d_t$-dimensional embedding for each mention using non-linear transformations.

\vspace{-.3cm}

\paragraph{Mouse trace encoder $e_m$} takes in the mouse traces for each mention parsed above after it is preprocessed into a 5D vector of coordinates and area, $(x_\text{min}, x_\text{max}, y_\text{min}, y_\text{max}, \text{area})$ \cite{meng2021connecting} and outputs a $d_m$-dimensional embedding. 
In \cite{changpinyo2021telling, pont2020connecting}, mouse trace embeddings have been exploited for image retrieval, however, we use them to resolve coreferences.
We concatenate each mention embedding extracted from $e_t$ with the mouse trace encoding $e_m$, denoted as $e_{tm}$ and apply additional nonlinear transformations (Joint encoder in \cref{fig:method-fig}) 
before feeding into the cross-attention module. 

\vspace{-.3cm}

\paragraph{Cross-attention module $a$} takes in the joint text-trace embeddings for all the words in each mention % ($\{e_t(w)\}\; \forall\; w \in m $) 
and returns a $d$-dimensional vector for each $m$ by taking a weighted average of them based on their correlations with the image regions.
% A naive way of representing a mention with $|m|$ words is to take the average of all the word (or joint word-trace) embeddings. 
Concretely, in this module, we first compute the correlation between each word $w$ (or joint word-mouse trace) and all regions, take the highest correlation over the regions through an auxiliary function $\bar{a}$:
\begin{equation}
    \label{eq.abar}
    \bar{a}(w) = \max_{r \in I} \big( \frac{ \exp(e_{tm}(w) \cdot e_i(r) )}{ \sum_{r' \in I} \exp(e_{tm}(w) \cdot e_i(r'))} \big)
\end{equation}
where $\cdot$ is dot product. 
The transformation can be interpreted as probability of word $w$ being present in image $I$.
Then we compute a weighted average of the word embeddings for each mention $m$: 
\begin{equation}
    \label{eq.a}
     a(m) = \sum_{w \in m} \bar{a}(w) e_{tm}(w).
\end{equation} 
Similarly, $a(m)$ can be seen as probability of mention $m$ being present in image $I$.

\vspace{-.3cm}

\paragraph{Scoring function $g(m,r)$} can be written as a dot product between the output of the attention module and region embeddings:
\begin{equation}
    \label{eq.g}
     g(m,r) = a(m) \cdot e_i(r).
\end{equation}
While taking a dot product between the two embeddings seemingly ignores the correlation between text and image data, the region embedding $e_i(r)$ encodes the semantic information about the detected object category in addition to other visual features and hence results in a high score only when the mention and region are semantically close.
Further implementation details about the modules can be found in \Cref{sec.exp} and the supplementary.

 \section{Experiments} 
\label{sec.exp}
% \subsection{Datasets}

% \noindent
% \textbf{Overview. }
% The main goal of our experiments is to test the efficacy of our proposed method on the problem of resolving multimodal coreferences. We also perform evaluation on weakly supervised visual phrase grounding .

% \noindent
% \textbf{Datasets. }
We train our models on the Flickr30k subset of the Localized Narratives~\cite{pont2020connecting} which consists of 30k image-narration pairs, and evaluate on the proposed \textbf{\OurDatasetShort} dataset, which contains 1000 and 800 pairs for test and validation respectively.
% For training, we have 30k images with paired narrations for each image.
% The evaluation dataset,  \OurDataset has 1000 images for test and 880 images for validation.

\vspace{.1cm}
\noindent\textbf{Evaluation metrics}.
% \subsection{Evaluation metrics}
% \noindent
% \textbf{Coreference evaluation.} 
To evaluate the CR performance, we use the standard link-based metrics MUC~\cite{vilain1995model} and BLANC~\cite{recasens2011blanc}:\footnote{Refer to \cite{lee2017end, luo2016evaluation} for a more detailed discussion of CR metrics.}
% A number of metrics have been proposed in the NLP literature to measure the task of coreference resolution \cite{lee2017end, luo2016evaluation}. In this paper, we use the link-based metrics MUC \cite{vilain1995model} and BLANC \cite{recasens2011blanc} to evaluate the coreference links predicted by our model. 
% We denote the predicted chains from the model as $R$ and the ground truth chains from human labeling as $K$. 
% For instance, let's say the ground-truth chain $K$ contains two chains with mentions as \textit{{a,b,c}} and \textit{{d,e,f,g}} and the output chain $R$ contains three chains with mentions \textit{{a,b}}, \textit{{c,d}} and \textit{{f,g,h,i}}. In this example, mention \textit{e} is missing from the response and mentions \textit{h} and \textit{i} are wrongly predicted. 
% Below, we explain how the two metrics compute precision and recall:

\noindent
\textbf{(a) {MUC F-measure}} counts the coreference links (pairs of mentions) common to the predicted chain $R$ and the ground-truth chain $K$ by computing MUC-R (recall) and MUC-P (precision). 

\noindent
\textbf{{(b) BLANC}} measures the precision (BLANC-P) and recall (BLANC-R) between the ground-truth and predicted coreference links and also between non-coreferent links. 
% \begin{equation}

%     R_n = \frac{|N_k \cup N_r|} {|N_k|}
% \end{equation}

% \begin{equation}
%     P_n = \frac{|N_k \cup N_r|} {|N_r|}
% \end{equation}

% \noindent
% \textbf{Narrative grounding evaluation.}
\noindent\textbf{{(c) Narrative grounding.}} For evaluating narrative grounding in images, we consider a prediction to be correct if the IoU (Intersection over Union) between the predicted bounding box and the ground truth box is larger than 0.5 \cite{wang2020maf, gupta2020contrastive}. 
We report percentage accuracy for evaluating narrative grounding for both noun phrases and pronouns. 
Further details about the metrics is in supplementary material.
% \parbox[t]{2mm}{\multirow{2}{*}{\rotatebox[origin=c]{90}{Language}}} &

\begin{table*}[!ht]
\renewcommand*{\arraystretch}{1.13}

		\resizebox{0.8\textwidth}{!}{
			\begin{tabular}{c|ccc|cccccc}

                 Method & Text  & Image   & MT     & MUC-R & MUC-P & MUC-F1 & BLANC-R & BLANC-P & BLANC-F1  \\
				\toprule
				% VLK \cite{lu2016visual} & 8.45 & - & 3.36  & 3.75 & - & - & 3.13 & 3.52 & - & -   \\
				% VTransE \cite{zhang2017visual} & - & -  & 2.65  & 3.51 & - & - &  1.71 & 2.14 & - & -   \\
				 Rule-Based \cite{lee2011stanford} & \cmark & \xmark & \xmark & 5.6 & 10.13  & 6.4 & 3.3 & 4.1  & 4.9 \\ 
			    Neural-Coref \cite{lee2017end}    & \cmark & \xmark & \xmark & 0.11 & 0.17 & 0.13 & 1.59 & 36.99  & 3.23 \\  
                 Similarity-based     & \cmark & \xmark & \xmark & 7.07 & 14.43 & 9.06  & 37.48  & 65.17  & 45.98 \\  \midrule
               GLIP \cite{li2022grounded}            & \cmark & \cmark & \xmark & 0.13 & 0.12 & 0.12 & 21.71& 61.40  & 31.66 \\
               MAF$^{\dagger}$ \cite{wang2020maf}            & \cmark & \cmark & \xmark & \textbf{25.86} & 10.18 & 13.21 & 37.68  & 61.14  & 38.17 \\      
                    
				% \multirow{5}{*}{Ours}           & \cmark & \cmark & \cmark & 24.36 & 17.55 & 18.60 & 67.46 & 68.08 & 66.00 \\
                                                % & \cmark & \cmark & \cmark & 33.78  & 20.42  & 23.30 & 71.22 & 69.55 & 67.73 \\  
				                                % & \cmark & \cmark & \cmark & 35.20 &  20.08 & 23.24  & 71.72  & 69.19 & 67.25  \\
                MAF++                              & \cmark & \cmark & \xmark & 19.07 & 15.62 & 15.65 & 41.25 & 65.04 & 47.21 \\ \midrule 
                \multirow{2}{*}{Ours}  & \cmark & \cmark  & \xmark & 22.07  & 17.10 & 17.58 & 42.72 & 65.92 & 48.29 \\
                                              & \cmark & \cmark & \cmark & {24.87} & \textbf{18.34} & \textbf{19.19}   &  \textbf{43.81} & \textbf{66.35}  & \textbf{48.53} \\
    			  
    		\bottomrule
			
\end{tabular}}
\caption{CR performance on \OurDatasetShort dataset. \textit{MT denotes mouse trace and $\dagger$ denotes our trained model.}}
\label{table:main_coref_results}
\end{table*}

% \subsection{Implementation details }
% \label{sec.imp}
\vspace{.1cm}
\noindent
\textbf{Inputs and modules.} For the image modeling, we extract bounding box regions, visual features and object class labels using the Faster-RCNN object detector \cite{ren2015faster}. 
For the text modeling, we use Glove embeddings \cite{pennington2014glove} to encode the object class labels and the mentions from the textual branch. 
For the mouse traces, we follow \cite{pont2020connecting} and extract the trace for each word in the sentence and then convert it into bounding box coordinates for the initial representation. 
The model discussed in \cref{sec.method} referred to as `Ours' in \cref{sec.results} uses the transformer backbone for the image, text and trace encoders (more details in supplementary). 

\vspace{.1cm}
\noindent
\textbf{Baselines. }
We consider the following baselines to fairly compare and evaluate our proposed method:

\noindent
\textbf{(a) Text-only CR: }For all these methods, we directly evaluate the coreference chains using the narration only without the image.  (1)~\textit{Rule-based \cite{lee2011stanford}:} In this method, a multi-sieve rule based system  is used to find mentions in the sentence and the coreference chains, 
(2)~\textit{Neural-Coref \cite{lee2017end}:} Instead of rules, this method is trained end-to-end using a neural network on a large corpus of wikipedia data to detect mentions and coreferences, and 
(3)~\textit{Similarity-based:} 
We compute cosine similarity between mentions using Glove word features and threshold them to get coreference chains.  

\noindent
\textbf{(b) Visual grounding: }The baselines discussed below are not trained for CR and hence we post-process their output in order to evaluate for CR.
(1)~\textit{GLIP \cite{li2022grounded}:} GLIP is trained on large-scale image-text paired data with bounding box annotations and shows improvement on object detection and visual phrase grounding. 
To evaluate it for CR, we predict bounding boxes for the mentions in the narrations from GLIP. 
If the IoU overlap between the mentions is greater than 0.7, then we consider them to form a coreference chain, 
(2)~\textit{MAF$^\dagger$ \cite{wang2020maf}:} MAF is a weakly supervised phrase grounding method, originally trained on the Flickr30k-Entities~\cite{plummer2015flickr30k}. 
We train this model on narrations data and evaluate CR by computing \cref{eq.f}.
(3)~\textit{MAF++: }We retrain the MAF$^{\dagger}$ model on the narrations with our regularization term. 
Architecturally our method differs from the MAF$^{\dagger}$ in two aspects: i) we employ a transformer to encode visual and text features unlike the MLP in theirs and ii) we attend to the mouse traces when present (not present in MAF) and word features jointly whereas they directly compute the similarity function.

\section{Results}
\label{sec.results}
\paragraph{Coreference resolution.} 

In \Cref{table:main_coref_results}, we report CR performance of the baselines and our method. 
% For the text-only baselines, we consider a rule-based method \cite{lee2011stanford} and an end-to-end trained neural CR  method~\cite{lee2017end} trained on large-scale supervised language coreference dataset. 
% For MAF, we train the model and report our results. 
% For our method, we report results with the transformer encoder (Tr) trained with the language prior regularization (Reg) in \Cref{sec.method}.
% Note that, unless described otherwise, our method corresponds to the variant that use the transformer variant, trained with image, text and mouse traces with the regularizer.
Our method significantly outperforms all the text-only and the grounding baselines on all the metrics.  
The text-only CR baselines in the first three rows fail to effectively resolve conferences from narrations. 
It is important to note that relatively high number in BLANC scores (compared to MUC) occure because this measures also counts non-coreferent links (\ie mentions that are not paired with anything), whereas MUC only measures pairs that are resolved. 

The rule-based method~\cite{lee2011stanford} uses exact match noun phrases, pronoun-noun matches and the distance between mentions as hard constraints. 
It achieves low scores on all metrics and especially on BLANC. The reason for this is the limitation of the rule-based heuristics:
For instance, in long narrations, if a pronoun such as \textit{she} occurs farther to its referent (\eg \textit{the woman}) than the predefined distance, it will not form a coreference chain.
In contrast, as we apply rules as a soft constraint, we are able to make more flexible decisions in our method. 
Neural-Coref~\cite{lee2017end}, a deep network on a pre-trained large-corpus of labeled CR data, obtains low scores on CIN for both MUC and BLANC. 
This is due to the large domain gap between the source and target data as well as the ambiguity in resolving the mentions without the visual cues. 
Similar observations are made when pretrained CR methods are applied to other domains such as biomedical text~\cite{lu2021coreference} or social media~\cite{aktacs2020adapting}.
Lastly, the similarity-based baseline performs poorly, as the utilized off-the-shelf word vectors are not trained to cluster corefering mentions. 
The relatively high scores on BLANC is due to the frequent non-coreferents in our narratives. 
This kind of approach clusters words with similar meaning together \eg \textit{woman} and \textit{another woman} (both representing female entities) or \textit{he} and \textit{she} (both pronouns).   

% Specifically, these models are trained to follow strict rules such as the pronominal pronouns mentioned later in the sentence are likely to be linked to the pronoun that is mentioned before (as also shown in \cref{fig:intro-fig}), hence the prior harms the performance. 
% This kind of prior is harmful in our setting. 
Next we compare our method to the visual grounding baselines that use both image and text input.
Our method also outperform these baselines:
% GLIP suffers a significant drop in performance in resolving coreferences. 
% More specifically, 
Though GLIP is pretrained on large-scale data with ground-truth boxes for each object in captions, these captions are usually short and do not contain multiple mentions of entities, unlike in our data.
Hence GLIP acts more like an object detector, fails to link coreferring pairs (low MUC scores) and merely identifies singletons (higher BLANC scores).
% The low performance of GLIP is due to it's inability to resolve instances of specific pronouns and nouns (\eg `a person' in \cref{fig:intro-fig}) could only correspond to one person and not all the people involved in the scene. 
% As this model is trained for object detection, it localizes all instances of a particular class. 
% Whereas, in our approach the soft constraint helps to suppress this behavior and hence learn stronger disambiguations. 
While it is nontrivial to finetune GLIP on our data without groundtruth boxes, we finetuned MAF on our data, as its training does not require groundtruth boxes; we denote this as MAF$^\dagger$.
This is the strongest baseline on our task, as training it on narrations including the pronouns reduces the domain gap and enables it to resolve coreferences well.
% MAF$^\dagger$ \cite{wang2020maf} retrained on image narrations achieves the strongest baseline performance. This is due to two reasons 1) training on narrations reduced the domain gap and 2) during training we also enforce the model to learn the grounding of pronouns.
However, this method obtains low precision by incorrectly linking visually similar mentions (that do not belong together) such as \textit{trees}, \textit{plant}, \textit{flowers}. 
When the training is regularized with the linguistic priors from our method, denoted as MAF++, its performance significantly improves on both MUC and BLANC. 
The constraint helps to push away the negative mentions (\textit{trees}, \textit{plant}, etc.) and encourage the model to learn unique embeddings for them. 
Due to the self-attention in the transformer architectures, Ours without mouse traces (MT) achieves better performance than MAF++, a simple MLP baseline.
% Although, it is still very challenging to learn such links (low precision scores) and hence, when trained with the constraint, MAF$^\dagger$++, it achieves much better F1 scores compared to the baseline.
The performance difference between our method without using mouse-traces and MAF++ can be explained by the better architecture described previously.
Finally, our method achieves the best performance gains in CR thanks to the mouse traces and improved architecture over MAF.
% which we ablate in \cref{sec.ablation}.

% Compared to the visual grounding base, our transformer encoder based method trained with the regularizer, with or without mouse traces, surpasses the performance of the baselines in all metrics. 
% The consistent gains in performance shows the effectiveness of our proposed method in resolving coreferences. 

% Note that the extremely low BLANC scores obtained by the text-only baselines is due to the fact that they only produce coreference chains (multi-mention entities) and therefore achieve lower scores, whereas our method also produces singletons (chains of length one). 
% In contrast, the MUC scores only measure coreference linking between pairs, so they are unaffected by singleton mentions. 

\noindent
\textbf{Ablation on mouse traces} In \Cref{table:main_coref_results}, we also analyze the contribution of modeling mouse traces (second last row). 
Adding the mouse traces improves performance on CR across all metrics. 
We hypothesize that the mouse traces provide a strong discriminative location prior to the textual mentions, which helps the model to learn a better compatibility score. 
To visualize qualitatively, consider the example in \Cref{fig:method-fig}, the same mention \textit{this person} points to two different visual regions -- one with the person holding the ball and the other person standing next to the door. In such cases, mouse traces provide a strong signal for disambiguation. 
But in many cases, mouse traces are noisy and can link mentions that are very close to each other in the image, referring to two different regions. 
In the above example, mouse traces for \textit{these persons} and \textit{this person} have a significant overlap and hence act as a noisy prior. 
Therefore, without the visual/image region features, it is very challenging to address the problem with mouse traces alone. 
% We can leverage the power of mouse traces with the self-attention in the \textit{joint text-trace encoder} transformer.
\vspace{-1.0em}

\begin{figure*}[t]
\begin{center}
\includegraphics[width=0.9\linewidth]{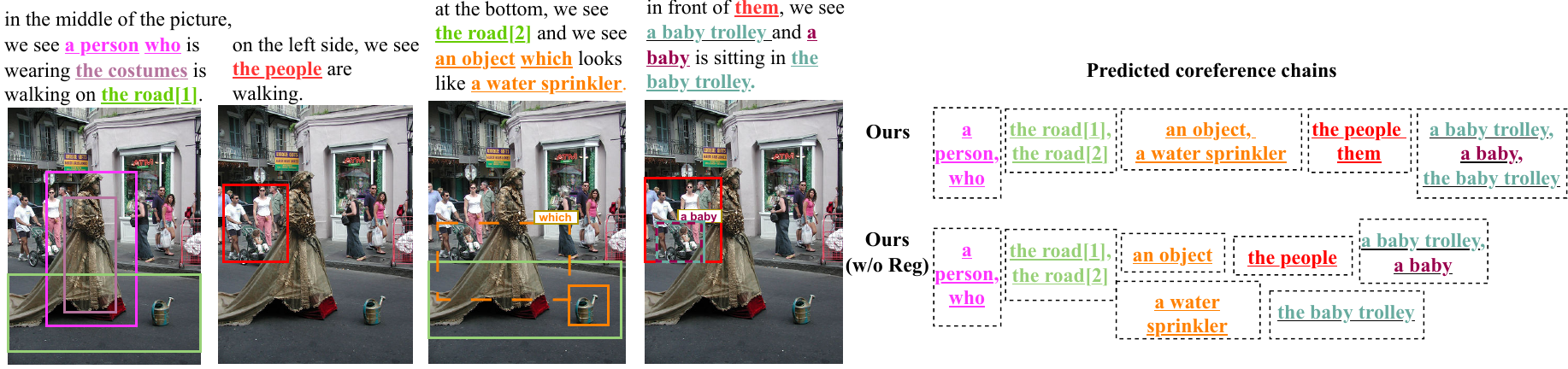}
\vspace{-0.5cm}
\end{center}
\caption{Qualitative results of predictions on the \OurDatasetShort dataset. The colored mentions in the text indicate the ground truth coreference chains. The solid and dotted bounding boxes on the image denote the correct and incorrect grounding respectively for our proposed method. We also show the predicted coreference chains for our final method with and without regularizer. }
\label{fig:qual_results}
\end{figure*}
% \vspace{-0.5em}

\paragraph{Narrative grounding} 
Not only does our method show performance gains on CR but also outperform the baselines on another challenging task of narrative grounding. 
\Cref{table:main_grounding_results} compares results from our methods and baselines. We directly compare with the weakly supervised method for a fair comparison.
MAF$^\dagger$~\cite{wang2020maf} is originally evaluated on the Flickr30k-Entities \cite{plummer2015flickr30k} dataset where the textual descriptions are significantly shorter (\ie single sentence) than the image narrations in our dataset. 
The performance of MAF on our dataset is significantly lower (21\% vs 61\% on Flickr30k-Entities), which indicates that narrative grounding is a challenge in itself and cannot be addressed off-the-shelf by phrase grounding methods. 
When trained with the regularizer, the localization performance improves for both nouns and pronouns with our method and MAF++. 
With the help of regularization, the model learns to attend to different regions of the image for semantically similar mentions as they might be two separate entities (\eg \textit{five people} and \textit{the people} in \cref{fig:qual_results}). 
% It is important to note that the grounding of noun/pronoun might be incorrect but it can still have the correct coreference pairs. For example, as shown on the right side in \cref{fig:qual_results}, \textit{some of them} and \textit{their} form the correct coreference pair but are not accurately grounded as depicted by the dotted blue lines. 

% \vspace{-1.0em}

\begin{table}[t]
\renewcommand*{\arraystretch}{1.13}

		\resizebox{0.65\textwidth}{!}{
			\begin{tabular}{c | c | c| c | c}

				 Method & Reg & Noun Phrases & Pronouns & Overall \\
     \toprule
        		
			    MAF$^\dagger$  \cite{wang2020maf} &\xmark & 21.60 & 18.31 & 20.91  \\
			
                     MAF++  & \cmark &  25.58 & 22.36  & 24.91   \\ 
                				\midrule
			    \multirow{2}{*}{Ours} & \xmark & 27.62 & 23.46 & 26.75  \\
				% & MLP & \cmark  & 27.73 & 24.66 &  27.09 \\
				& \cmark & \textbf{30.27}  & \textbf{25.96}  &  \textbf{29.36} \\
				
    		\bottomrule
    		
\end{tabular}}
\caption{Grounding accuracy (\%) for noun phrases and pronouns and the overall accuracy on the \OurDatasetShort dataset.}
\label{table:main_grounding_results}
\end{table}

% Our method significantly outperforms the baseline on grounding accuracy for both the noun phrase mentions and pronouns. Hence, we are not only able to resolve coreferences but also ground the mentions to the image with a higher level of contextual reasoning.

% Compared to the performance on Flickr30k Entities dataset (reported as 61\%) where the textual descriptions are significantly shorter, the performance of even the strongest baseline method is significantly lower on our \OurDataset dataset. 
% This clearly shows that grounding is much more difficult when we are dealing with long-form descriptions rather than single sentences. 

% \begin{table}[!ht]
% \renewcommand*{\arraystretch}{1.13}

% 		\resizebox{0.9\textwidth}{!}{
% 			\begin{tabular}{c | c| c | c| c | c}

% 				 Method & Arch & Reg & Noun Phrases & Pronouns & Overall \\
%      \toprule
        		
% 			    MAF$^\dagger$  \cite{wang2020maf} & & & 21.60 & 18.31 & 20.91  \\
% 				\midrule
% 			    \multirow{4}{*}{Ours} & MLP & \xmark  & 25.58 & 22.36  & 24.91  \\
% 			    & Tr & \xmark & 27.62 & 23.46 & 26.75  \\
% 				& MLP & \cmark  & 27.73 & 24.66 &  27.09 \\
% 				& Tr & \cmark & \textbf{30.27}  & \textbf{25.96}  &  \textbf{29.36} \\
				
%     		\bottomrule
    		
% \end{tabular}}
% \caption{Grounding accuracy (\%) for noun phrases and pronouns and the overall accuracy on the \OurDataset dataset.}
% \label{table:main_grounding_results}
% \end{table}
% \vspace{-0.5em}

\begin{table}[!h]
\renewcommand*{\arraystretch}{1.13}

		\resizebox{0.6\textwidth}{!}{
			\begin{tabular}{c | c  c | c }

				&  \multicolumn{2}{c|}{CR} & Grounding\\
            				
				 Attention Type &   MUC-F1  & BLANC-F1 &  Acc(\%)   \\ \toprule
				  {Average}  & 17.02 & 48.26  & 28.83 \\

		           {Cross attention}  & \textbf{19.19} &\textbf{48.53} & \textbf{29.36} \\

    		\bottomrule
			
\end{tabular}}
\caption{Our method with/without cross attention.}
\label{table:ablation}
\end{table}
\vspace{-1.5em}

\vspace{-0.5em}

\paragraph{Further ablations.}
\label{sec.ablation}
\Cref{table:ablation} compares the performance of our final method under two settings: (1)~directly averaging the word features or (2)~attending over the words by using the image as the query as discussed in \cref{sec.method}. Both the narrative grounding accuracy and the coreference evaluation get a boost in performance for visually aware word features. More often than not, the word phrases are relatively short (\eg,~\textit{the machine}) and hence the model does not always learn to disambiguate better with attention for the grounding. On the other hand, this technique is especially useful for CR because the flow of visual information to the word features acts as a prior to cluster mentions that refer to the same region but with are referred to with different mentions/entities in the text (\eg \textit{the machine } and \textit{an equipment}). We provide detailed ablations in the supplementary material. 

\paragraph{Qualitative results }
\Cref{fig:qual_results} qualitatively analyzes CR and narrative grounding.
% trained with the combined loss function or only with contrastive loss. 
We visualize the narrative grounding results from our proposed method on the images. The model correctly resolves and localizes phrases such as \textit{a person, who}, \textit{the people, them} and \textit{a baby trolley, the baby trolley}. Whereas, the model fails to ground and chain the instance \textit{\textcolor{maroon}{a baby}}. It is interesting to note that our model pairs \textit{an object} and \textit{water sprinkler}, thereby resolving ambiguity in what \textit{the object} might refer to. But it fails to add \textit{\textcolor{orange}{which}} to this coreference chain. 
Moreover, without the language regularizer, our method fails to link \textit{\textcolor{red}{them}} to \textit{the people}. 
It is very hard to learn coreferences for these pronouns as they come with a weak language prior and hence are difficult for the model to disambiguate. 
Our model (without regularization) misses the referring expression of \textit{\textcolor{bluegreen}{the baby trolley}} to refer to the instance of the trolley before. With the help of rules (\eg last token match), we can resolve these pairs more often than not. 
Hence, we clearly show the challenging problem of coreferences we are dealing with and indicate the great potential for developing models with strong contextual reasoning.

\section{Conclusion}
\label{sec.conclusion}
We introduced a novel task of resolving coreferences in image narrations, clustering mention pairs referring to the same entity.
% Existing multi-modal datasets have largely ignored the task of CR, clustering mention pairs referring to the same entity.
For benchmarking and enabling the progress, we introduce a dataset -- \OurDatasetShort -- that contains images with narrations annotated with coreference chains and their grounding in the images. 
We formulate the problem of learning CR by using weak supervision from image-text pairs to disambiguate coreference chains and  linguistic priors to avoid learning grammatically wrong chains. 
We demonstrate strong experimental results in multiple settings. 
In the future, we plan to address the noise induced by the language rules during learning and also reduce the errors coming from the mouse traces.
We hope that our proposed task definition, dataset and the weakly supervised method will advance the research in multi-modal understanding.

% \noindent
% \textbf{Future Direction and Limitations. }

% \hb{any limitations?}

%%%%%%%%% REFERENCES
{\small
\bibliographystyle{ieee_fullname}
\bibliography{egbib}
}

\newpage
\section*{Appendix}

\section{Annotation Details}

\paragraph{Localized Narratives dataset. }
Tuset \etal \cite{pont2020connecting} proposed the Localized Narratives dataset, new form of multimodal image annotations connecting vision and langauge. In particular, the annotators describe an image with their voice while simultaneously hovering their mouse over the region they are describing. Hence, each image is described with a natural language description attending to different regions of the image. In addition to textual descriptions (obtained using speech to text conversion), they additionally provide mouse traces for the words.

The Localized Narratives dataset is built on top of COCO \cite{lin2014microsoft}, Flickr30k \cite{plummer2015flickr30k}, ADE20k \cite{zhou2017scene} and Open Images \cite{kuznetsova2020open}. The statistics of the individual datasets are shown in \Cref{table:dataset_stats}. 

\begin{table}[!ht]
\renewcommand*{\arraystretch}{1.13}

		\resizebox{0.8\textwidth}{!}{
			\begin{tabular}{c | c| c | c }
				\hline
				
				   Localized Narratives Subsets \cite{pont2020connecting} & \#images & \#captions & \#words/capt.   \\
				\hline

			    COCO & 123,287  & 142,845  & 41.8\\
                Flickr30k  & 31,783 & 32,578 & 57.1 \\
                ADE20k & 22,210 & 22,529 & 43.0 \\
                Open Images & 671,469 & 675,155  & 34.2\\
    			  
    		\hline
    			  
\end{tabular}}
\caption{Statistics of Localized Narratives for COCO,
Flickr30k, ADE20k, and Open Images.}
\label{table:dataset_stats}
\end{table}

\begin{figure*}[!ht]
\begin{center}
\includegraphics[width=\linewidth]{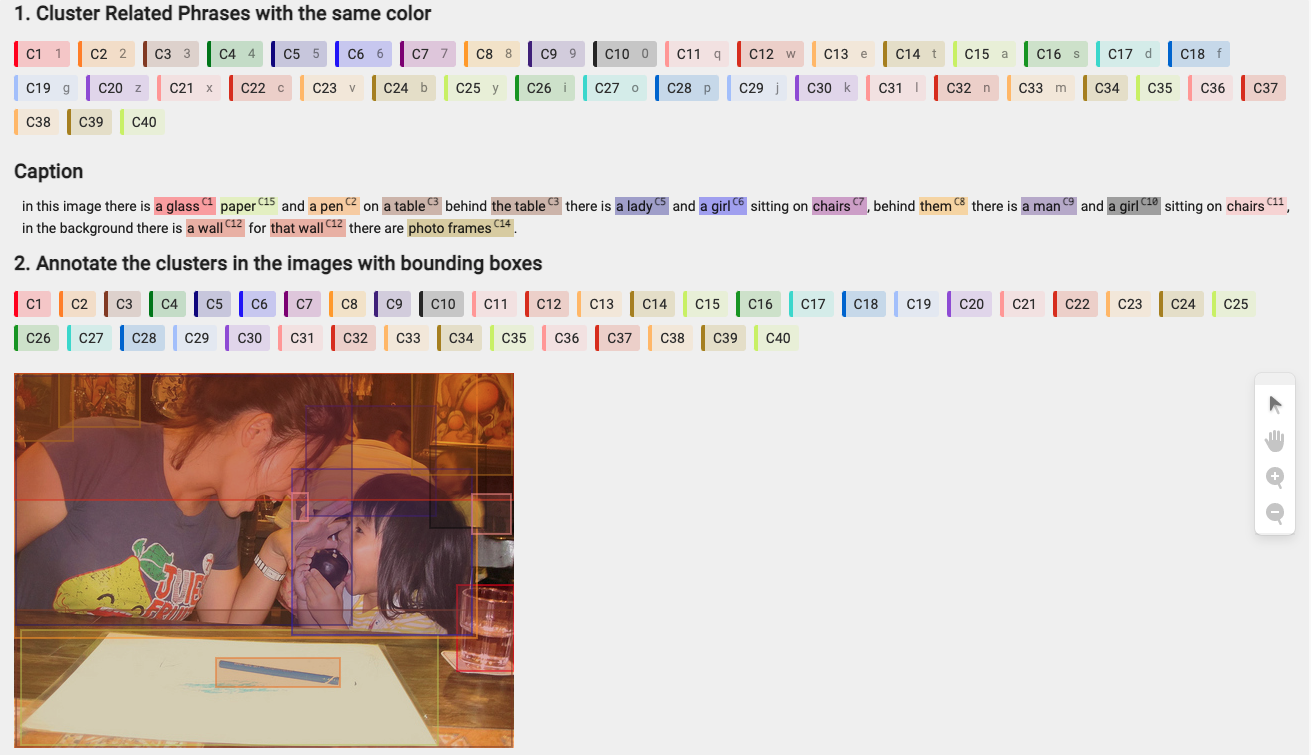}
% \vspace{-0.5cm}
\end{center}
\caption{Annotation interface from Label Studio.}
\label{fig:annotation_tool}
\end{figure*}

\paragraph{Annotation tool and analysis. }
We develop an HTML based interface on the Label Studio annotation tool \cite{labelstudio}. \Cref{fig:annotation_tool} shows the annotation interface from Label Studio. We hired 6 high quality annotators (all from computer science background) for an average of 54 hours of annotation time. The annotators were trained with the exact description of the task and given a pilot study before proceeding with the complete annotations. The pilot study was useful to correct and retrain annotators if needed. 
As shown in \Cref{fig:annotation_tool}, the annotators had to select a mention in the caption with a given label (C1, C2, etc.) in Step 1 and draw a bounding box in the image for the selected mention in Step 2 (with the same label).

For Step 1, if the mention is coreferring then it is selected with the same label to define coreference chains. It is important to note that the captions are pre-marked with noun phrases parsed from \cite{spacy}. The annotators are instructed to correct the phrases if they are wrong (\eg for a mention glass windows, the parser parses \textit{glass} and \textit{windows} as two different mentions rather than belonging to the same label/cluster) and remove the phrases that do not correspond to region in the image.  

In Step 2, if there are plural mentions such as \textit{two men}, we ask the annotators to draw two separate bounding boxes for this. In the case of mentions such as \textit{several people} if the people are less than five, they are instructed to draw separate bounding boxes otherwise a group bounding box (covering all the people). 

Given the challenging nature of the task, we doubly annotate 30 images with coreference chains and bounding boxes to compute the inter-annotator agreement. More specifically, for the coreference chain we compute \textit{Exact Match} which denotes whether the coreference chains annotated by the two annotators are the same. We get an exact match of 79.9\% in the coreference chains, which is a high agreement given the complexity of the task. For the bounding box localization, we compute the Intersection over Union (IoU) to compute the overlap between the two annotations. It is considered to be correct/matching if the IoU is above 0.6. We achieve bounding box accuracy of 81\% on this subset of images. This analysis shows good agreement between the annotators given the subjective nature and complexity of the task.

\paragraph{\OurDataset dataset. }

In total, we annotate all the 1000 test images and 880 validation images (out of 1000) in the Flickr30k dataset. The text descriptions from the Localized Narratives dataset are very noisy with a lot of words/sequence of words. We manually filter phrases such as - \textit{in this image, in the front, in the bakcground, we can see, i can see, in this picture}. If there are some other mentions that are pre-marked and not filtered, we ask the annotators explicitly to filter them out. By doing this, we make sure that the dataset is clear of any unnecessary and noisy mentions. 

All the words that are marked as mentions and are not noun phrases (as detected by the part of speech tagger \cite{spacy}) are considered as pronouns \eg \textit{them, they, their, this, that, which, those, it, who, he, she, her, him, its}. 

\begin{figure}[!ht]
\begin{center}
\includegraphics[width=\linewidth]{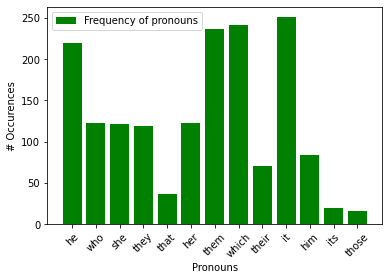}
% \vspace{-0.5cm}
\end{center}
\caption{Total number of occurrences of pronouns in \OurDataset.}
% \textit{Figure Source: Localized Narratives \cite{pont2020connecting}.}}
\label{fig:pronoun-freq}
\end{figure}

\begin{figure}[!ht]
\begin{center}
\includegraphics[width=\linewidth]{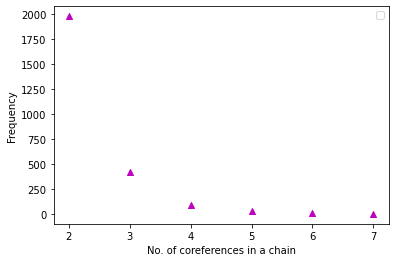}
% \vspace{-0.5cm}
\end{center}
\caption{Number of coreference chains with 2 or more than 2 mentions in a chain in \OurDataset.}
% \textit{Figure Source: Localized Narratives \cite{pont2020connecting}.}}
\label{fig:chain-freq}
\end{figure}

\begin{table*}[t]
\centering
\renewcommand*{\arraystretch}{1.13}

		\resizebox{0.8\textwidth}{!}{
			\begin{tabular}{cc|cccccc | c}

		     & & \multicolumn{6}{c|}{CR} & Grounding \\
				\toprule
				
                  MT & Loss Function    & MUC-R & MUC-P & MUC-F1 & BLANC-R & BLANC-P & BLANC-F1  & Acc (\%)\\
				\toprule
                \xmark & \xmark & 21.84  & 12.29 & 14.09  & 40.15 & 62.82  & 43.69 & 25.97  \\
                \cmark & \xmark & 20.19 & 15.79 & 16.26  & 41.91 & 65.42  & 47.82 & 26.75 \\ \midrule
                  \cmark & L1 &20.76  & 15.47 & 16.05   & 41.73  & 64.94  & 47.09 & 27.65 \\

                 \cmark &  MSE & 21.58 & 16.40 &  17.00  & 42.19  & 65.37 & 47.60 & 28.50 \\ \midrule
                \xmark & \multirow{2}{*}{Frobenius Norm}  & 22.07  & 17.10 & 17.58 & 42.72 & 65.92 & 48.29 & 28.31 \\
                \cmark &  & \textbf{24.87} & \textbf{18.34} & \textbf{19.19}   &  \textbf{43.81} & \textbf{66.35}  & \textbf{48.53} & 29.36\\
    			  
    		\bottomrule

\end{tabular}}
\caption{Ablation study with different regularizer types and without mouse traces.}
\label{table.extra_ablation}
\end{table*}

\paragraph{Statistics for the \OurDataset. }

In \Cref{fig:pronoun-freq}, we show the statistics for the frequency of pronouns in the dataset. Few pronouns (\eg he, it, them) are more frequent than the others. Overall, the occurence of pronouns is frequent to conduct a fair evaluation of the coreference based models. 
Similarly in \Cref{fig:chain-freq}, we evaluate how many mentions occur in the coreference chains. Coreference chains with 2 and 3 mentions have a very high frequency in the dataset. There are few chains that have longer mentions (\eg 6 and 7). Hence, we can safely conclude that the dataset is a powerful tool to evaluate coreference chains and learn complex coreferencing and grounding models.
Moreover, the average length of the mentions (excluding pronouns) is 1.93.

\section{Evaluation Metrics}    
In this section, we discuss in detail the evaluation metrics used for CR and narrative grounding.
For CR, we use the MUC and the BLANC metrics, which are discussed below.

\noindent
\textit{(a) {MUC F-measure.}} It measures the number of coreference links (pairs of mentions) common to the predicted $R$ and ground-truth chains $K$. 
It involves computing the partitions with respect to the two chains:
\begin{equation}
    \text{MUC-R} = \frac{\sum_{i=1}^{N_k} (|K_i| - |p(K_i)|)}{\sum_{i=1}^{N_k} (|K_i| - 1)},
\end{equation}
\begin{equation}
    \text{MUC-P} = \frac{\sum_{i=1}^{N_r} (|R_i| - |p^{'}(R_i)|)}{\sum_{i=1}^{N_r} (|R_i| - 1)}
\end{equation} 
where $K_i$ is the $i^{th}$ ground-truth chain and $p(K_i)$ is the set of partitions created by intersecting $K_i$ with the output chains; $R_i$ is the $i^{th}$ output chain and $p^{'}(R_i)$ is the set of partitions created by intersecting $R_i$ with the ground-truth chains; and $N_k$ and $N_r$ are the total number of ground-truth and output chains, respectively.

\noindent
\textit{{(b) BLANC.}} Let $C_k$ and $C_r$ be the pairs of coreference links respectively, and $N_k$ and $N_r$ be the set of non-coreference links in the ground-truth and output respectively. The BLANC Precision and Recall for coreference links is calculated as follows: 

$R_c = \frac{|C_k \cup C_r|} {|C_k|}$ and $P_c = \frac{|C_k \cup C_r|} {|C_r|}$, where $R_c$ and $P_c$ are the recall and precision respectively. 

Similarly, recall $R_n$ and precision $P_n$ for non-coreference links ($N_k$ and $N_r$) are computed.  
The overall precision and recall are: 

$\text{BLANC-R} = \frac{(R_c + R_n)} {2}$ and $\text{BLANC-P} =\frac{(P_c + P_n)} {2}$, respectively.

 For evaluating narrative grounding in images, we consider a prediction to be correct if the IoU (Intersection over Union) score between the predicted bounding box and the ground truth box is larger than 0.5 \cite{wang2020maf, gupta2020contrastive}.  Following \cite{kamath2021mdetr}, if there are phrases with multiple ground truth boxes (\eg several people), we use the any-box protocol \ie, if any ground truth bounding box overlaps the predicted bounding box, it is a correct prediction. We report percentage accuracy for evaluating narrative grounding.

% Frequency of pronouns
% {'he': 220, 'who': 122, 'she': 121, 'they': 119, 'that': 37, 'her': 122, 'them': 237, 'which': 241, 'their': 70, 'it': 251, 'him': 84, 'its': 19, 'those': 16}

% average length of mentions - 1.923061214070557

% length of clusters - {3: 425, 2: 1982, 4: 96, 5: 32, 6: 10, 7: 2}

% \begin{table}[!ht]
% \renewcommand*{\arraystretch}{1.13}

% 		\resizebox{0.8\textwidth}{!}{
% 			\begin{tabular}{c | c| c}
% 				\hline
% 				&  \multicolumn{2}{c}{Phrase Grounding Accuracy} \\
				
% 				 Method  & Original Flickr30k & CRG - Flickr30k (Ours)   \\
% 				\hline

% 				\textbf{Weakly Supervised} & & \\
%                 MAF \cite{} & & \\
%                 Image-Text Transformer (Ours) & & \\

%     		\hline

% \end{tabular}}
% \caption{Performance of Weakly Supervised Models on the Original Flickr30K and our CRG - Flickr30k dataset.}
% \label{table:baseline1}
% \end{table}

% \begin{figure}[!ht]
% \begin{center}
% \includegraphics[width=0.6\linewidth]{figures/mouse_fig.drawio.pdf}
% % \vspace{-0.5cm}
% \end{center}
% \caption{Localization of the textual spans in the sentence description. The localization is obtained by converting the trace strokes to bounding boxes following \cite{pont2020connecting}.}
% \label{fig:mouse-fig}
% \end{figure}

\section{Implementation details }
\label{sec.imp}
\noindent
\textbf{Inputs and modules.} For the image modeling, we extract bounding box regions, visual features and object class labels using the Faster-RCNN object detector \cite{ren2015faster}. 
We use Glove embeddings \cite{pennington2014glove} to encode the object class labels and the mentions from the textual branch. 
For the mouse traces, we follow \cite{pont2020connecting} and extract the trace for each word in the sentence and then convert it into bounding box coordinates for the initial representation. 
% There are two variations of our model, one with a two-layer MLP and another with a transformer encoder. 
All the modules \ie, image encoder, text encoder, trace encoder and joint text-trace encoder are a stack of two transformer encoder layers. 
Each transformer encoder layer includes a multi-head self attention layer and an FFN. 
There are two heads in the multi-head attention layer, and two FC layers followed by ReLU activation layers in the FFN. The output channel dimensions of these two FC layers are 2048 and 1024, respectively. 
The input to the joint text-trace encoder comes from the separate text and trace encoder branches. We add a special embedding to the learned embeddings following \cite{chen2020uniter} to distinguish between the two modalities (text and trace) in the transformer encoder. 

\noindent
\textbf{Training details.} The whole architecture is trained end-to-end with the AdamW \cite{loshchilov2017decoupled} optimizer. 
% For training the MLP, the learning rate is set to $\text{1e-4}$, batch size is set to 64, weight decay is 1e-4 and the loss coefficient $\lambda$ is set to 0.01. 
We train the transformer encoders with the learning rate of $\text{3e-5}$, batch size of eight, weight decay of 0.01 and the loss coefficient $\lambda$ of 0.001. 
We train the model for 60 epochs and choose the best performing model based on the validation set.

\section{Ablation Study}

% \noindent
% \textbf{Impact of regularizer.}

In \Cref{table.extra_ablation}, first we study the impact of training with just our proposed architecture without the mouse traces and regularizer. The model suffers a drop in both the CR and grounding performance. While the model is able to learn some coreference links but it still produces a lot of false positives (lower precision scores), compared to the model trained with mouse traces.  
Next, we study the effect of training with different regularizer types. We achieve improvement in performance with the Frobenius norm as a constraint unlike L1 and MSE, as it imposes a stronger constraint on the learned coreference matrix. 
Note that the last row corresponds to our proposed model (MT+Frobenius norm).

\begin{figure*}[!ht]
\begin{center}
\includegraphics[width=\linewidth]{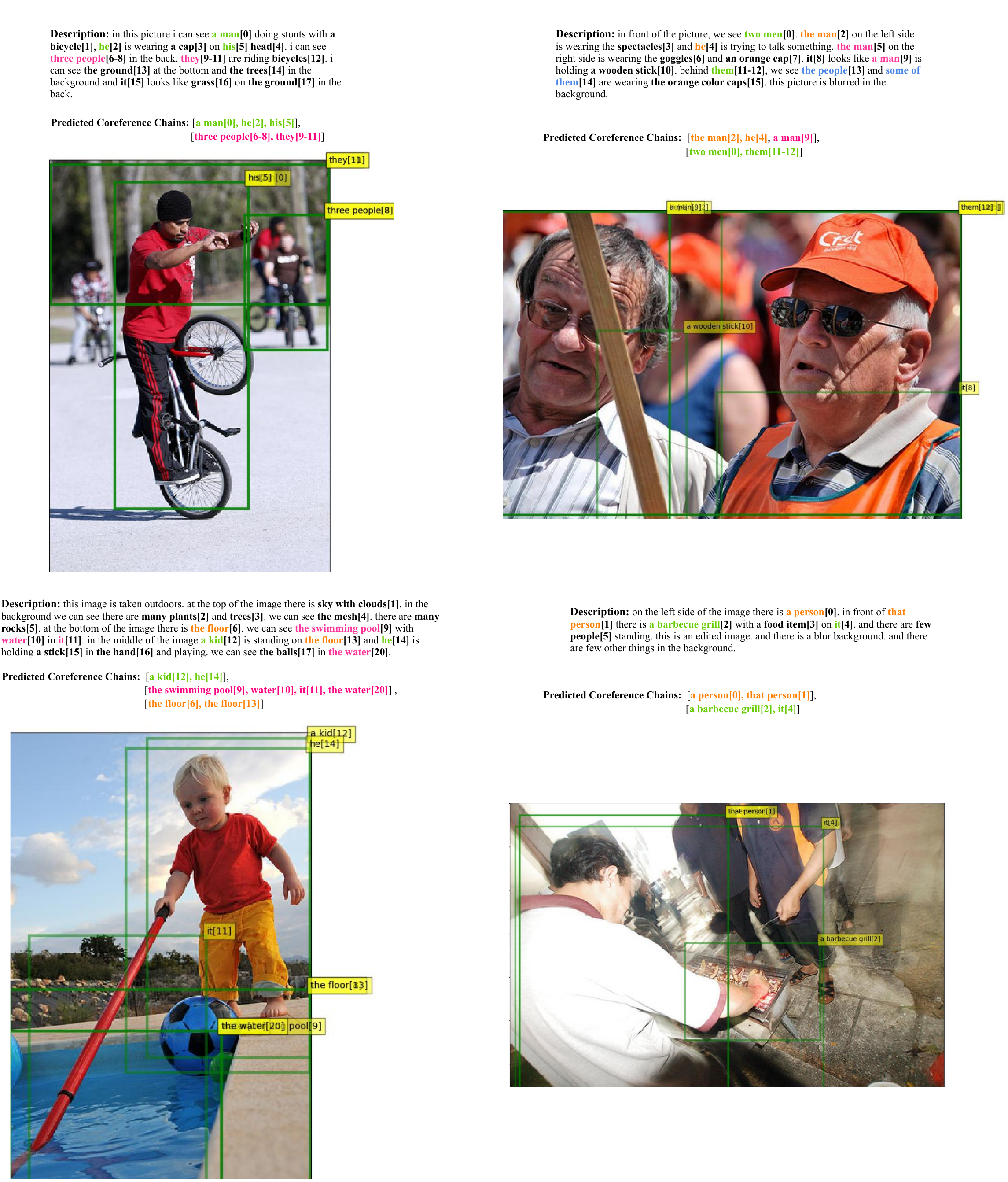}
% \vspace{-0.5cm}
\end{center}
\caption{Additional qualitative results for coreference chains. For each image, we show the predicted coreference chain (mentions more than 2) and the grounding results for the corresponding mentions in the chain. The colored mentions in the descriptions are the ground-truth coreference chains. }
\label{fig:qual-vis}
\end{figure*}

\section{Additional Qualitative Results}

In \cref{fig:qual-vis}, we show additional qualitative results from our proposed method. The model correctly chains mentions and grounds them to the correct entities in the image even for complex and ambiguous cases. Our model finds coreferences for people (\eg \textit{[a man, his]}) or for objects (\eg \textit{[a barbecue grill, it]}). Moreover, it also finds links for plurals such as \textit{[two men, them]}. There is a huge potential in learning to disambiguate the mentions in the descriptions and this work paves the way for future research.

% \section{Future Directions}       

\end{document}